\documentclass{article}

    \PassOptionsToPackage{compress}{natbib}    

\usepackage[final]{neurips_2023}

\usepackage{blindtext}
\usepackage{caption}
\usepackage[utf8]{inputenc} 
\usepackage[T1]{fontenc}    
\usepackage{hyperref}       
\usepackage{rotating}
\usepackage{url}            
\usepackage{booktabs}       
\usepackage{amsfonts}       
\usepackage{nicefrac}       
\usepackage{microtype}      
\usepackage{xcolor}         
\usepackage{graphicx} 
\usepackage{wrapfig}
\usepackage{multirow}
\usepackage{adjustbox} 
\usepackage{amsmath}
\usepackage{wrapfig,lipsum,booktabs}

\usepackage{float}

\title{Exploring Attribute Variations in Style-based GANs using Diffusion Models}

\author{%
  Rishubh Parihar \\
  Indian Institute of Science, Bangalore\\
  \texttt{rishubhp@iisc.ac.in} \\
  \And
  Prasanna Balaji \\
  Indian Institute of Science, Bangalore\\
  \texttt{bprasanna@iisc.ac.in} \\
  \And
  Raghav Magazine \\
  Indian Institute of Technology, Dharwad\\
  \texttt{200010042@iitdh.ac.in} \\
  \And
  Sarthak Vora \\
  UCLA\\
  \texttt{sarthakvora@smail.itm.ac.in} \\
  \And
  Tejan Karmali \\
  Avataar AI\\
  \texttt{tejank10@gmail.com} \\
  \And
  Varun Jampani \\
  Stability AI \\
  \texttt{varunjampani@gmail.com} \\
  \And
  R. Venkatesh Babu \\
  Indian Institute of Science, Bangalore\\
  Karnataka, India \\
  \texttt{venky@iisc.ac.in} \\
}

\begin{document}

\maketitle

\begin{abstract}
Existing attribute editing methods treat semantic attributes as binary, resulting in a single edit per attribute. However, attributes such as eyeglasses, smiles, or hairstyles exhibit a vast range of diversity. In this work, we formulate the task of \textit{diverse attribute editing} by modeling the multidimensional nature of attribute edits. This enables users to generate multiple plausible edits per attribute. We capitalize on disentangled latent spaces of pretrained GANs and train a Denoising Diffusion Probabilistic Model (DDPM) to learn the latent distribution for diverse edits. Specifically, we train DDPM over a dataset of edit latent directions obtained by embedding image pairs with a single attribute change. This leads to latent subspaces that enable diverse attribute editing. Applying diffusion in the highly compressed latent space allows us to model rich distributions of edits within limited computational resources. Through extensive qualitative and quantitative experiments conducted across a range of datasets, we demonstrate the effectiveness of our approach for diverse attribute editing. We also showcase the results of our method applied for 3D editing of various face attributes. 
\end{abstract}

\section{Introduction}
\vspace{-3mm}
The recent emergence of deep generative models employing GANs ~\cite{GAN-goodfellow, karras2020analyzing, eg3D} has significantly transformed image generation and editing. Various methods have been proposed that leverage disentangled latent space of GANs for attribute editing in 2D~\cite{shen2020interpreting,harkonen2020ganspace,patashnik2021styleclip,abdal2021styleflow} and in 3D~\cite{vive3d,preim3d}. Existing methods treat semantic attributes as binary and are limited to generating a single edit for a given attribute. However, most \textit{attributes are multidimensional} in nature, e.g., smiles, eyeglasses, and hairstyles. To this end, we explore a new perspective for attribute editing and propose to learn the distribution over plausible attribute edits. This enables users to generate multiple edit variations for a given attribute and select the best one. E.g., it can allow a user to generate diverse eyeglass styles and identify the most fitting option. In this work, we formulate the task of \textit{diverse attribute editing} and propose a method to generate multiple attribute edit variations. 

\begin{wrapfigure}{r}{0.45\linewidth}
    \vspace{-8mm}
    \includegraphics[width=0.9\linewidth]{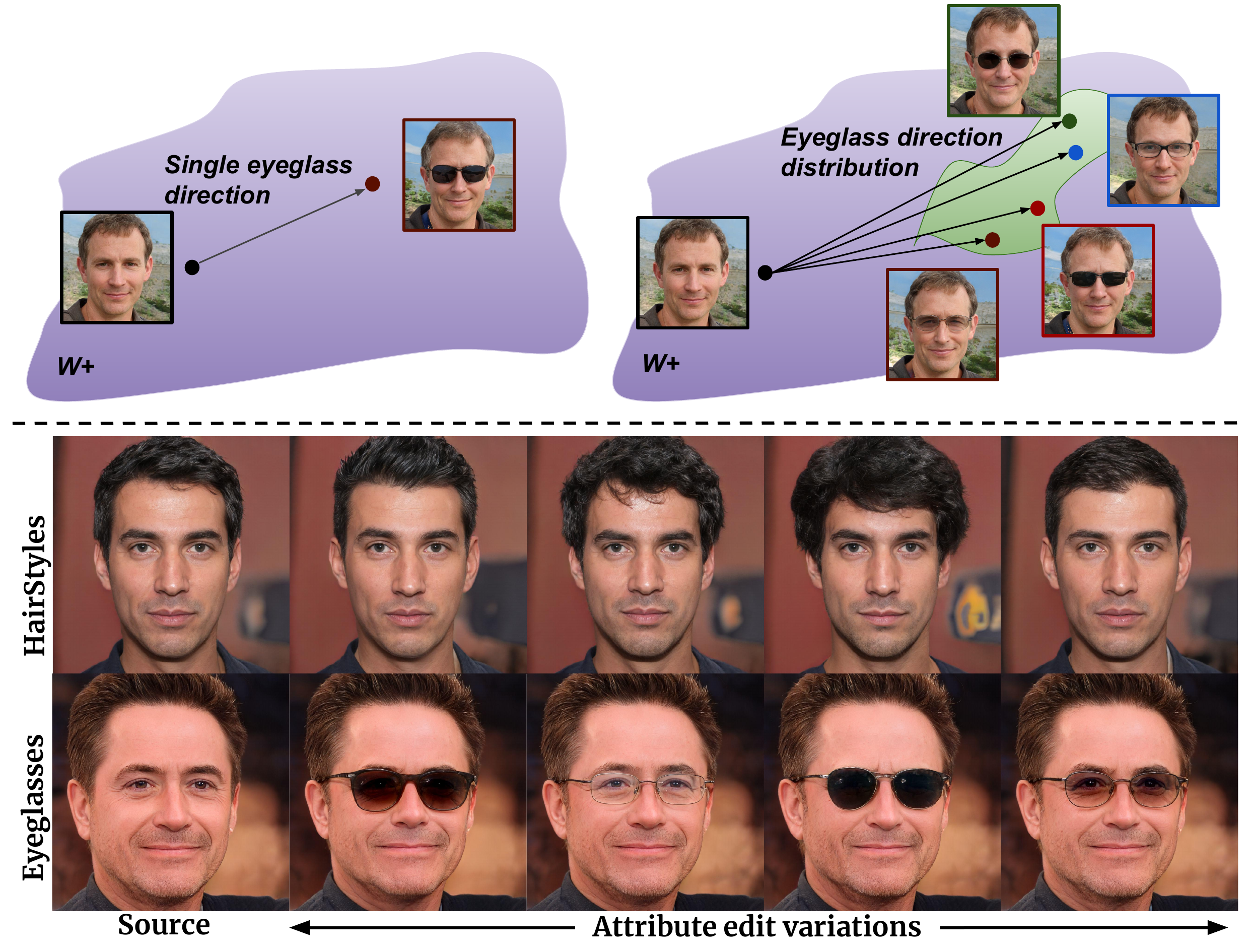}
    \caption{Existing methods consider attributes as binary and obtain a single attribute edit direction, our proposed method generates a distribution over edit directions, allowing for multiple attribute variations}
    \label{fig:teaser}  
    \vspace{-4mm}
\end{wrapfigure}

Style-based GAN models have semantically rich and disentangled $\mathcal{W/W+}$ latent spaces. This is proven by the existence of linear latent directions in StyleGANs that control a single attribute~\cite{shen2020interpreting,sefa} in the generated image. Leveraging this, current attribute editing methods~\cite{patashnik2021styleclip, shen2020interpreting, abdal2021styleflow, harkonen2020ganspace} treat attributes as unidirectional and discover a single linear direction in the latent space to modify these attributes. 
For instance, they obtain a single direction to add a smile or eyeglasses. However, the representation of diverse edits for smiles or eyeglasses using a single edit direction is restrictive, given the wide range of variations in these attributes. 
Motivated by this, we propose to train a generative model over the distribution of edit directions for each attribute (Fig.~\ref{fig:teaser}).

As the distribution of attributes is extremely rich and contain a large range of variations in appearance and structure, eg., eyeglasses have variations in frames shapes, lens shape, material and colors. To this end, we propose to used diffusion model (implemented as a small MLP) in the latent space due to their excellent mode coverage capabilities. 
The motivation to apply diffusion in GAN latent space is two folds: 1) \textit{it enables us to exploit attribute disentanglement properties in the space for diverse attribute editing;} 2) \textit{ diffusion model training and inference in the compressed latent space is computationally inexpensive.} More specifically, we acquire a dataset of edit directions in the compressed latent space space for training a DDPM~\cite{ddpm} model.

In summary, our contributions are as follows: 
\vspace{-3mm}
\begin{enumerate}
\item {A novel method to learn the space of diverse attribute variations with Diffusion Model in the latent space of pretrained style-based GANs.} 
\item {State-of-the-art results in generating diverse attribute edit variations on multiple attributes and datasets and comprehensive analysis of different model components.}
\item {Generalization of the proposed method for editing in 3D and out-of-domain images}.
\end{enumerate}


    

\section{Methodology}
\begin{figure*} 
    \centering    \includegraphics[width=\linewidth]{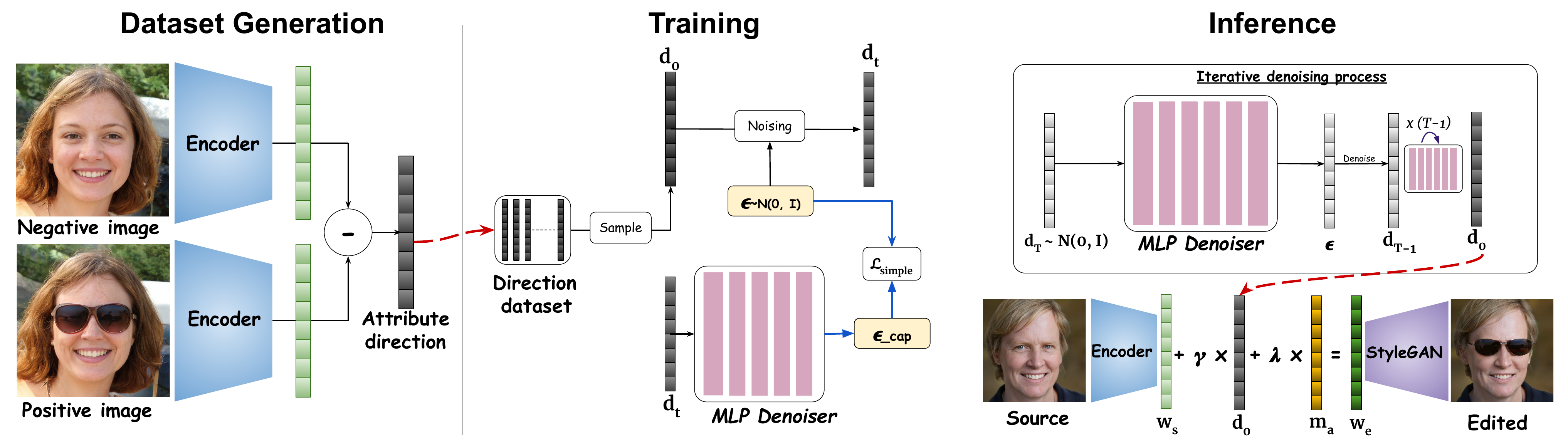}
    \caption{Our methodology for diverse attribute editing comprises three major stages: 1) \textbf{Dataset Generation.} We create a dataset of edit directions by embedding negative and positive image pairs into the latent space and computing the difference between these directions. 2) \textbf{Training.} We train a DDPM model over the dataset of edit directions for the given attribute employing a denoising objective. 3) \textbf{Inference.} To edit a new image, we first encode it into the latent space and then add an edit direction sampled with iterative denoising in the reverse diffusion process.}
    \label{fig:2_method}
    \vspace{-4mm}
\end{figure*}

\vspace{-3mm}
\noindent \textbf{Approach Overview}. The proposed method for diverse attribute variations includes three main steps (illustrated in Fig.~\ref{fig:2_method}). Firstly, we generate a dataset of image pairs, each with a single attribute change, and derive a set of edit directions from the difference between corresponding latent codes (Sec.~\ref{sec:data_gen}). Secondly, we use the dataset of edit directions to train a DDPM~\cite{ddpm} model to capture the space of diverse attribute edits (Sec.~\ref{sec:model-training}). Lastly, during inference, we obtain a new edit direction from the trained model and combine it with the source latent code to produce an edited image (Sec.~\ref{sec:attr-edit}). We provide a detailed explanation of each of these steps in the following sections. In our experiments, we explore $\mathcal{W/W+}$, latent spaces of the style-based GANs for obtaining editing directions.

\vspace{-3mm}
\subsection{Data Generation}
\label{sec:data_gen}
\vspace{-3mm}
We generate a synthetic dataset of image pairs to obtain a dataset of disentangled edit directions for a given attribute $a$. These image pairs consist of a positive and a negative image, where the positive image $I^a_p$ has the attribute $a$, and the negative image $I^a_n$ does not. We create these image pairs such that all the other attributes and identity are unchanged in $I^a_p$ and $I^a_n$. To obtain an edit direction $d_a$, we first encode the image pairs into the $\mathcal{W+}$ latent space using e4e~\cite{e4e} encoder model $\mathcal{E}$ and take a difference between them. As the image pairs have modifications in only a single attribute, the difference vector $d_a$ captures transformation corresponding to only attribute $a$.  
\vspace{-1mm}
\begin{equation}
\label{eq:edit_dir}
    d_a = \mathcal{E}(I^a_p) - \mathcal{E}(I^a_n)
\vspace{-3mm}
\end{equation}

We make a dataset $\mathcal{D}_a = \{d_a^i\}$ of $N$ edit directions for attribute $a$ by accumulating these edit directions. To obtain the dataset of disentangled image pairs with a single attribute change, we use off-the-shelf attribute editing methods ~\cite{patashnik2021styleclip,sam_age} for hairstyles and age attributes. For eyeglass and smile attributes, we use the method proposed in ~\cite{parihar2022everything}. Further details about the dataset creation are provided in the supplementary document. We generate a dataset of image pairs for each attribute with $30K$ source images from CelebA-HQ~\cite{karras2017progressive} dataset. Next, we encode the image pairs and obtain the edit direction using Eq.~\ref{eq:edit_dir}.

\vspace{-3mm}
\subsection{Diffusion model training}
\label{sec:model-training}
\vspace{-3mm}
Diffusion models (DM) being likelihood models, are shown to learn complex image distributions with excellent mode coverage ability in the data distribution.  Latent DMs~\cite{ldm} train DM on features and we go further by modeling over the $\mathcal{W+}$ space of StyleGAN models as an image representation, because of its high compression. We train a DDPM on the $\mathcal{W+}$ space to model the multimodality in the attribute variations.


Specifically, given a dataset of diverse attribute edit directions $\mathcal{D}_a$, we train a DDPM model to learn the distribution over edit directions. The goal is to sample a new edit direction from the DDPM model that models a novel and realistic transformation only in the attribute $a$. 
During training, we randomly sample an edit direction $d$ from $\mathcal{D}_a$, and corrupt it with a gaussian noise \begin{math}\boldsymbol{\epsilon}\end{math} $\sim$ \begin{math}\mathcal{N}(0,I)\end{math}. Following the convention, we denote the selected sample as $\mathbf{d_0}$: $\mathbf{d_t} = \sqrt{\bar{\alpha}_t} \mathbf{d_0} + \sqrt{1-\bar{\alpha}_t}\epsilon$ with $\bar{\alpha} = \prod_{i=1}^T\alpha_i$, and $0 = \alpha_T < \alpha_{T-1} < ..... < \alpha_0 = 1$, being hyperparameter of diffusion schedule. We implement the denoiser network $\epsilon_\theta(\boldsymbol{d_t}, t)$
as a time-conditioned Multi-Layer Perceptron (MLP) network. To train the denoising network, we use the simple loss~\cite{ddpm}, between added noise $\epsilon$ and ${\epsilon_\theta(d_t,t)}$ as $\mathcal{L}_{simple} = \mathbb{E}_{\mathbf{d_0}, t, \boldsymbol{\epsilon}}[\|\boldsymbol{\epsilon} - \epsilon_\theta(\mathbf{d_t}, t)\|^2_2]$. As a normalization step, we subtract the mean direction $\mathbf{m_a}$ of the dataset $\mathcal{D}_a$ from the vectors $\mathbf{d_a}$ and normalize them to unit length before training. 




\vspace{-3mm}
\subsection{Diverse attribute editing}
\label{sec:attr-edit}
\vspace{-3mm}
Given a DDPM model on edit directions, we show how to use it to edit a given source image $I_s$ diversely. To perform diverse attribute editing, we first map it to its corresponding latent code $\mathbf{w_s}$, where $\mathbf{w_s} = \mathcal{E}(I_s)$. To obtain a new edit attribute direction $\mathbf{d_0}$, we first sample a noised direction $\mathbf{d_T}\sim\mathcal{N}(0, I)$, and iteratively denoise it using trained MLP denoiser $\epsilon_\theta$. Finally, we scale and shift the sampled latent direction ($\mathbf{d_0}$) as $\mathbf{d'} = \gamma * \mathbf{d_0} + \lambda * \mathbf{m_a}$, before finally adding it to the source latent code ($\mathbf{w_s}$) as $\mathbf{w_e} = \mathbf{w_s} + \mathbf{d'}$. The edited latent code $\mathbf{w_e}$ is then passed through the pre-trained StyleGAN2 model $\mathcal{G}$ to obtain the edited image $I_e$, where $I_e=\mathcal{G}(\mathbf{w_e})$, diversity parameter $\gamma$ and scale parameter $\lambda$ are the hyperparameters, and $\mathbf{m_a}$ is the mean edit direction for attribute $a$. We find that the hyperparameter $\gamma$ controls the diversity in the edits and $\lambda$ controls the strength of the edit, as supported by the analysis in the supp. mat..


\vspace{-3mm}
\section{Experiments}
\vspace{-3mm}
\noindent
We provide extensive experiments and ablations to evaluate our model for diverse attribute editing. First, we explain the datasets used, followed by results on diverse attribute editing for faces. Further, we present results for attribute editing on out-of-domain face images from Metfaces~\cite{Metfaces}, cars, and church datasets~\cite{lsun}. Finally, we present results for 3D-aware attribute editing on EG3D~\cite{eg3D}. Please check the supplementary for more visual results.

\begin{figure}[h]
    \centering    
    \includegraphics[width=1.0\linewidth]{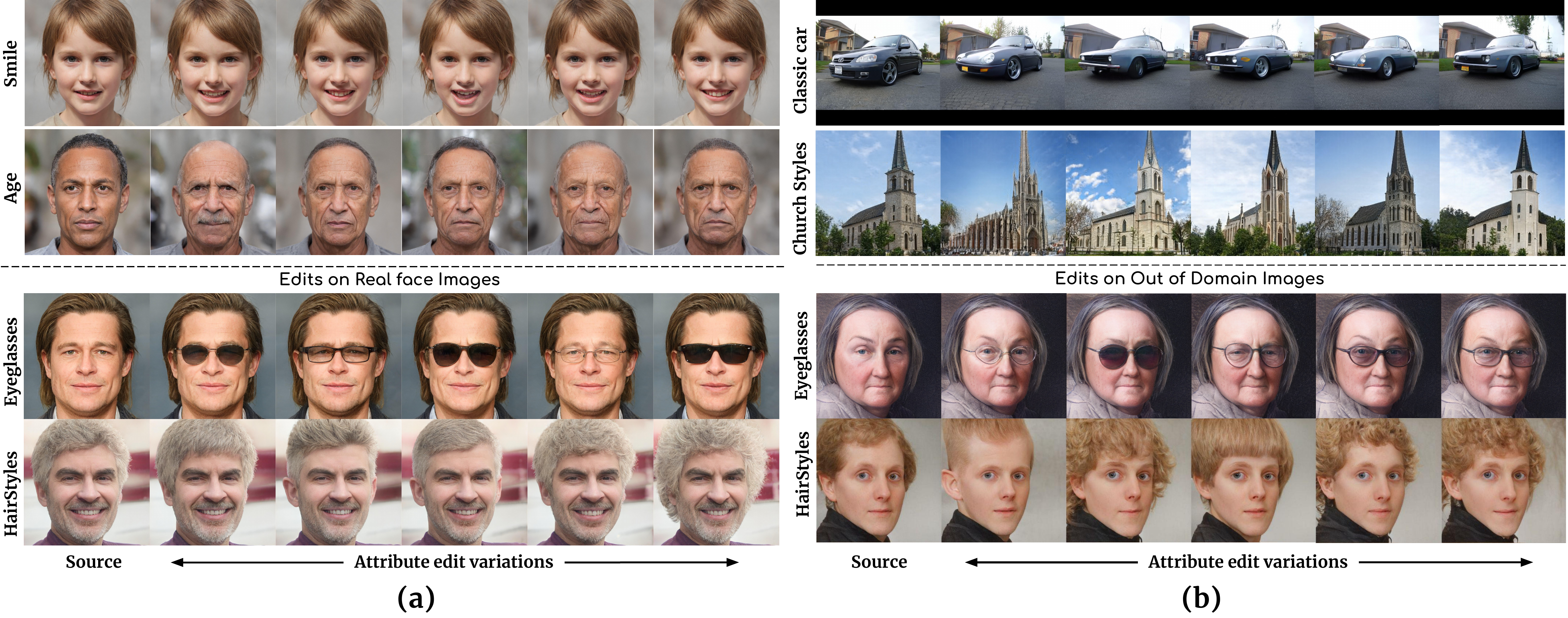}
    \vspace{-5mm}
    \caption{\textbf{a)} Diverse attribute editing on (Top) synthetic faces and (Bottom) real face images. The model generates diverse edit directions that are disentangled and preserve the subjects' identity. \textbf{b)} Diverse attribute variations for \textit{classic car} and \textit{church styles} (Top) and diverse attribute editing on out-of-domain painting images from Metfaces.}
    \label{fig:diverse_attr_edits_combined}
    \vspace{-6mm}
\end{figure}

\vspace{-3mm}
\subsection{Diverse attribute editing}
\vspace{-2mm}
\noindent
We present results for hairstyle, smile, eyeglass, and age attribute variations generated by our method in Fig.~\ref{fig:diverse_attr_edits_combined}. Our method generates different hairstyles - bangs, mohawks, curls, and short hairs while retaining other features. Similarly, our method can generate diverse smile and age variations in a disentangled manner with identity preservation. Our proposed method can generate diverse eyeglasses with variations in frame shapes, sizes, and colors of frames. Observe that all the edit variations preserve the subject's identity and other attributes. Our method generates diverse edits on real images as well (Fig.~\ref{fig:diverse_attr_edits_combined} Bottom).

\begin{wrapfigure}{o}{0.45\linewidth}
    \vspace{-2mm}
    \includegraphics[width=1.0\linewidth]{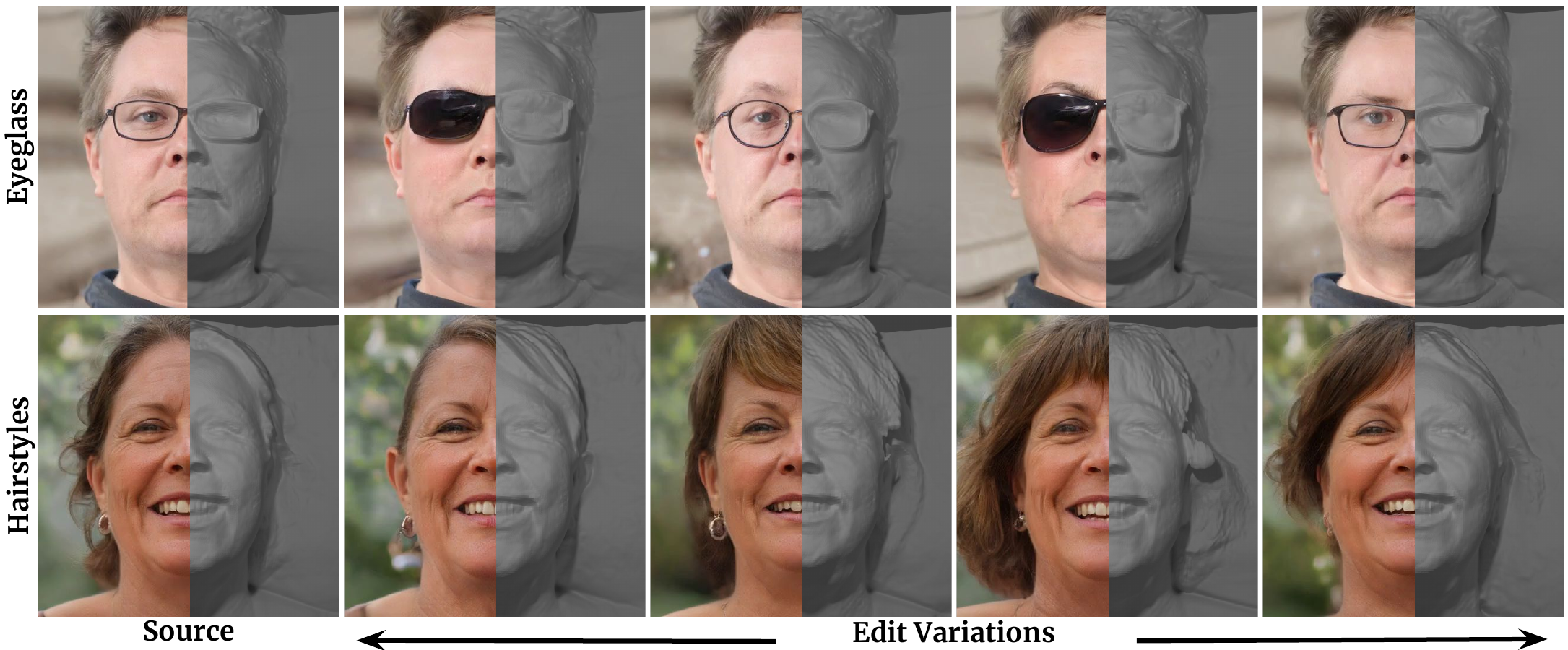}
    \caption{Diverse edits for eyeglass attribute on EG3D in 3D consistent manner. Observe the modified geometry of the eyeglass frames in both outputs.}
    \vspace{-2mm}
    \label{fig:14_eg3d}  
\end{wrapfigure} 


\textbf{Quantitative comparison.} We generate five edits for each attribute for a synthetic test set of $1000$ images to evaluate the quality and diversity of the edits. We compute FID, cosine similarity between face embeddings ~\cite{curr-face-rec} (CS), improved precision (P), and recall metrics (R)~\cite{kynkaanniemi2019improved}. We compare our method against - 1) Baseline - random edit directions sampled from the training dataset and 2) FLAME~\cite{parihar2022everything}, which is a few-shot method and performs diverse edits. We used the smaller set of $50$ edit directions from our training set and obtained edit directions from the FLAME method. Results are shown in Tab. ~\ref{tab:quant_compare}. We note that the proposed method performed best in identity preservation and visual quality measured by FID. In most cases, it obtained the highest precision and recall suggesting diversity in the output generations. The superior performance of our method to baseline in identity preservation suggests that the trained DDPM model is robust to outliers.

\begin{wraptable}{r}{0.45\textwidth}
    \vspace{-2mm}
    \caption{Quantitative comparison for diverse attribute editing.}
    \vspace{-2mm}
    \begin{adjustbox}{width=\linewidth}
    \label{tab:quant_compare}
    \begin{tabular}{c|c|ccccc}
    \toprule
    Attr.                  & Method   & CS $\uparrow$   & ED $\downarrow$ & FID $\downarrow$ & Prec. $\uparrow$ & Recall $\uparrow$ \\ 
                               \hline
    \multirow{4}{*}{Hairstyle} & Baseline & 0.869 & 0.79 & \underline{41.50} &   0.62&    0.92 \\
                               & FLAME    & \underline{0.956} & \underline{0.45} &44.87 &  \underline{0.79} &   \underline{0.97} \\
                               & Ours     & \textbf{0.973} & \textbf{0.35} &\textbf{39.58} &  \textbf{0.91}&   \textbf{0.98}\\ 
                               \hline
    \multirow{4}{*}{Eyeglass}  & Baseline & 0.931 & 0.53 & 70.91 &  \underline{0.66 }&  \underline{0.28} \\
                               & FLAME    & \underline{0.948} & \underline{0.49} &  \underline{69.50}&  0.61 &   0.22  \\
                               & Ours     & \textbf{0.958} & \textbf{0.41} & \textbf{66.65}  &\textbf{0.69} &    \textbf{0.35} \\ 
                               \hline
    \multirow{3}{*}{Smile}     & Baseline & 0.920 & 0.55 & 54.44 &  \underline{0.74} &  0.09  \\
                               & FLAME    & \underline{0.953} & \underline{0.48}  & \underline{53.29} & 0.73 &   \textbf{0.16}  \\
                               & Ours     & \textbf{0.969} & \textbf{0.43}  & \textbf{49.51}&  \textbf{0.77}&   \underline{0.14} \\ 
    \bottomrule
    \end{tabular} 
    \end{adjustbox}
\end{wraptable}

\textbf{Generalization and OOD Results.} We present results on cars, churches, and out-of-domain painting images from Metfaces~\cite{Metfaces} in Fig.~\ref{fig:diverse_attr_edits_combined}.
We generate diverse types of "classic cars" with high fidelity. For churches, we change the style of the church keeping while preserving the outer structure. For Metfaces, we generate multiple attribute edit directions from our diffusion models trained with real image pairs as explained in Sec~\ref{sec:model-training}. We can observe that the generated directions generalize well to the out-of-domain painting images and generate diverse attribute edits. Notably, the styles of the generated edits blend seamlessly with the painting styles.

\textbf{Editing on 3D aware GANs.} Our method generalizes for diverse attribute edits on 3D aware generative model EG3D~\cite{eg3D}.
We present the geometry of the edited outputs, where we can clearly observe shape changes associated with eyeglass edits.

\section{Conclusion} 
\noindent 
This work explores the challenging problem of diverse attribute editing by latent space manipulation in pre trained style-based GANs. Deviating from the existing method that estimates a single edit direction for a given attribute, we learn a diffusion model over the edit directions to learn the multimodal nature of the edits. Extensive results show that the proposed method generates diverse edits for real images, out-of-distribution images and 3D edits. The limitation of our method is reliance on synthetic image pairs to train the model. As a future work, text-driven multimodal editing can be explored. Additionally, a fine-grained control for multimodal attribute editing is interesting. 

\clearpage

\bibliographystyle{unsrtnat}
\bibliography{neuripsw}

\end{document}


\maketitle

\appendix
\renewcommand*\contentsname{Organization of Appendix}
\tableofcontents
    

\section{Related works}
\label{sec:related works} \textbf{Latent space-based image editing.}
Recently, various image and video editing works have been proposed that leverage semantics in the StyleGAN's latent space ~\cite{shen2020interpreting} to edit images~\cite{jahanian2019steerability, harkonen2020ganspace, abdal2021styleflow, patashnik2021styleclip, parihar2022everything, tewari2020stylerig, stylespace, parihar2023we, tzaban2022stitch}. One direction of works obtains a global edit direction in the latent space for each attribute~\cite{shen2020interpreting, harkonen2020ganspace, jahanian2019steerability, spingarn2020gan}. Traversal along these global directions edits the corresponding attribute in the generated image. Another cohort of methods obtains a local direction for each latent code. Essentially a non-linear mapping is learned between the input latent code and the desired edit code, using transformer networks ~\cite{transformereditsg2, umetani2022user, xu2022transeditor}, a mapper network  ~\cite{patashnik2021styleclip}. To obtain the edit direction for a given attribute, these works use - attribute classifiers~\cite{abdal2021styleflow, liang2021ssflow}, segmentation masks ~\cite{ling2021editgan}, clip-supervision ~\cite{patashnik2021styleclip,abdal2022clip2stylegan, zheng2022bridging} or perform unsupervised decomposition of the latent space ~\cite{jahanian2019steerability, harkonen2020ganspace,sefa,latentclr}. StyleFlow~\cite{abdal2021styleflow} learns a conditional normalizing flow network to learn a deterministic mapping from a source latent to a single edited latent code for each attribute. To enable editing on real images, encoder-decoder frameworks have been proposed that map the real image into the $\mathcal{W+}$ space and use StyleGAN's generator as the frozen decoder after latent editing~\cite{e4e,abdal2019image2stylegan,alaluf2021hyperstyle,psp}.


\textbf{Non-binary attribute editing.} Some editing works try to model the continuous attribute variations instead of treating attributes as binary. Works like ~\cite{patashnik2021styleclip, yang2021beyondbinary} obtain directions for non-binary attributes such as image style or expression change, but they also use a single edit direction for the given attribute/style. Unsupervised methods ~\cite{latentclr, harkonen2020ganspace} can learn a finite set of disentangled directions controlling each attribute. StyleSpace~\cite{stylespace} learns a set of disentangled vectors in style space controlling each attribute. However, these methods learn only a finite set of edit directions and cannot cover all the possible variations for any given attribute. FLAME~\cite{parihar2022everything} proposed a task of attribute style manipulation, where they generate edit variations by navigating in the attribute manifold. Furthermore, StyleFusion~\cite{kafri2021stylefusion} showed that a pre-trained StyleGAN could be used to decompose spatial semantic regions. In contrast to these approaches, we learn a distribution over edit directions for a given attribute and sample multiple variations of an edit. 

\textbf{Diffusion Models} Diffusion Models (DM) are likelihood-based models that have achieved state-of-the-art performance in sample generation~\cite{ldm} and density estimation~\cite{diffusion-beats-gan}. In contrast to GANs, DMs being likelihood-based models, prevent mode collapse and learn rich multi-modal distributions. DMs modeled as hierarchical denoising autoencoders ~\cite{sohl2015deep, ddpm} are trained to iteratively denoise images starting from pure noise. Due to the sequential nature of DMs, applying them in the pixel space for high-resolution images leads to high training costs and slow inference speeds~\cite{diffusion-beats-gan}. To this end, latent diffusion models (LDMs)~\cite{ldm} have been proposed that first encode the images into much lower dimensional spatial latent codes and apply DM in the latent space. Subsequently, multiple works are proposed that perform latent space diffusion for motion synthesis~\cite{diffusion-motion}, language generation~\cite{diffusion-language-generation}, point clouds generation~\cite{diffusion-point-cloud}, generating brain imaging ~\cite{diffusion-brain-imaging}. In contrast, we compress the image into semantic $\mathcal{W+}$ space of StyleGAN2 instead of spatial latent space and perform diffusion process in this highly compressed space.

\begin{figure}[h]
    \centering    \includegraphics[width=1.0\linewidth]{figs/asm-iccv-supply-fig2-compare-baseline.pdf}
    \caption{(Top) Comparison with the baseline edit directions - We randomly sample five edit directions each from $\mathcal{D}_a$ and our method for each attribute. Observe that using the baseline directions changes the subject's identity and other attributes. (Bottom) To visualize the disentanglement of attributes, we perform editing on a source image with $1000$ editing vectors from our method and baseline method and take the pixel-wise difference between the source and the edited image. We plot the per-pixel standard deviation of the difference image for eyeglass and hairstyle variations. Observe that for eyeglass edit, our method changes only the region near the eye, but the baseline method changes other regions such as hair and mouth. Similarly, for hairstyles, our method performs changes in localized hair regions.}
    \label{fig:s2_qual_compare_baseline}
\end{figure}

\section{Analysis of diverse edit directions} 
\label{sec:baseline_compare}
\noindent 
\textbf{Comparison with baseline edit directions.} In summary, our proposed method first creates a dataset $\mathcal{D}_a$ of edit directions using synthetic image pairs (Sec. {3.1} in the main paper). Later, a DDPM model is trained over these directions that can generate novel and diverse edit directions (Sec. {3.2} in main paper). A natural question is why we cannot directly use the edit directions from $\mathcal{D}_a$ to perform attribute editing (Baseline). In other words, is the edit direction space learned by the DDPM actually better than having the dataset of edit directions? Since the image pairs are synthetically generated, some of these edit directions are not clean and entangle multiple attributes and change the subject's identity as illustrated in Fig.~\ref{fig:s2_qual_compare_baseline}(Top). However, the proposed method can generate diverse edits while preserving the identity and attributes of the subject. 

To further analyze the disentanglement properties, we sample $1000$ edit directions from $\mathcal{D}_a$ and our trained diffusion model to perform edits on an input source image. We plot per pixel standard deviation of the $1000$ different images between the source and edit for both methods in Fig. ~\ref{fig:s2_qual_compare_baseline}(Bottom). We observe that the edits performed by our method change a localized region of interest in the edited image, whereas the baseline edits unnecessarily alter other regions. This analysis shows the better disentanglement properties of the directions generated by our method compared to baseline directions. Although our model is trained on the dataset $\matcal{D_a}$ of baseline edit directions, it can generate disentangled and identity-preserving edits. This indicates that the diffusion model is learning the space defined by the most dominant edit directions and is robust to outlier directions. 

To quantitatively compare the identity preservation ability of our method against the baseline, we performed the following analysis. We plot the Euclidean Distance (ED) between face embeddings (extracted from ~\cite{face-rec-2020}) of the source and edited image for the two methods in Fig.~\ref{fig:s1_identity_preservation}. As indicated by lower ED scores, our method has better identity preservation ability than the baseline. 

\begin{figure}[h]
    \centering    \includegraphics[width=1.0\linewidth]{figs/asm-iccv-supply-fig1-elu-plots.pdf}
    \caption{\textbf{Better identity preservation.} Distribution of the Euclidean Distance (ED) between the face recognition embeddings of edited and the source image over $1000$ images. Observe that our directions have lower ED demonstrating better identity preservation.}
    \label{fig:s1_identity_preservation}
\end{figure}  

\section{Generalization of edit directions.} 
To analyze the generalization capability of the edit directions for different editing, we perform editing with a single direction on multiple source images. Specifically, we sample a set of edit directions for from a trained diffusion model for hairstyles and eyeglasses and perform edits on six source images as shown in Fig.~\ref{fig:4_multi_dirs}. Editing with a direction generates similar styles in all the source images. For example, for hairstyles, $d_1$ generates curly hair, $d_2$ generates bangs and $d_3$ generates mohawk hairstyle. Observe that for each eyeglass direction, similar frame shapes are generated. For eyeglasses, $d_1$ generates ellipse-shaped frames, $d_3$ generates similarly shaped yet thicker frames and $d2$ generates round-shaped frames.

\begin{figure}[h]
    \centering    
    \includegraphics[width=1.0\linewidth]{figs/asm-iccv-fig4-multi-dimensional-edits.pdf}
    \caption{Results for editing multiple source images with the same direction. Each row, other than the source row, results from editing six different source images using the same edit direction $d_i$, sampled using the appropriate DDPM model. We present results for hairstyle edits on top and eyeglass edits at the bottom. Observe that for each edit direction, an attribute edit of a similar style is generated for all the source images.}
    \label{fig:4_multi_dirs}
\end{figure} 

\section{Comparison with multi-direction based editing methods}
\label{sec:multi_compare}
We compare our method for multi-dimensional edits with two methods that can generate a set of few directions for each attribute - StyleSpace~\cite{stylespace} and LatentCLR~\cite{latentclr}. StyleSpace learns the transformation over the style space of StyleGAN2 to edit any attribute. In their work, it has shown that there are few $(3-10)$ style codes that can control a single attribute such as hairstyle or smile. However, for some attributes, e.g., age, there is no disentangled style code. LatentCLR proposed a contrastive learning approach to find a finite set of edit directions in the $\mathcal{W}$ latent space, which contains complementary information. The edit directions for any given attribute need to be manually selected by carefully inspecting the learned set of directions. Both methods have some potential to perform more than one edit per attribute but are limited to only a finite number of edits. 

We compare our method against these methods for hairstyle attribute, as we find maximum directions for hairstyles in both of these methods. Specifically, we found $8$ directions from StyleSpace (($3-169$, $3-379$, $6-83$, $6-175$, $6-285$, $6-364$, $12-266$, $12-330$) as reported in ~\cite{stylespace} and $6$ directions for LatentCLR (when trained with non-linear method). Note that our method can sample a large number of directions as we have learned a generative model over edits. To obtain more variation in the edits from StyleSpace and LatentCLR, we randomly sampled the edit strength from a range of values working well for the respective methods. We present our results in Fig.~\ref{fig:s5_qual_comparision_multi_dir}. Our method generates diverse variations with minimal entanglement. StyleSpace can generate a few hairstyle variations, but most of the generated hairstyles are highly similar (similar hairstyles are highlighted using the same colored boxes). For LatentCLR, most of the obtained hairstyle directions entangle skin tone, lighting, and other attributes. Also, for LatentCLR, we observe that most of the hairstyles have similar structures but have variations in hair length and color. However, our method can generate a range of diverse hairstyle edits such as bangs, mohawk, curly, short hair, and further variation among those. This supports the need for a multi-direction editing method that can model the distribution over attribute edits that are beyond finite directions per attribute.

\begin{figure}
    \centering    
    \includegraphics[width=1.0\linewidth]{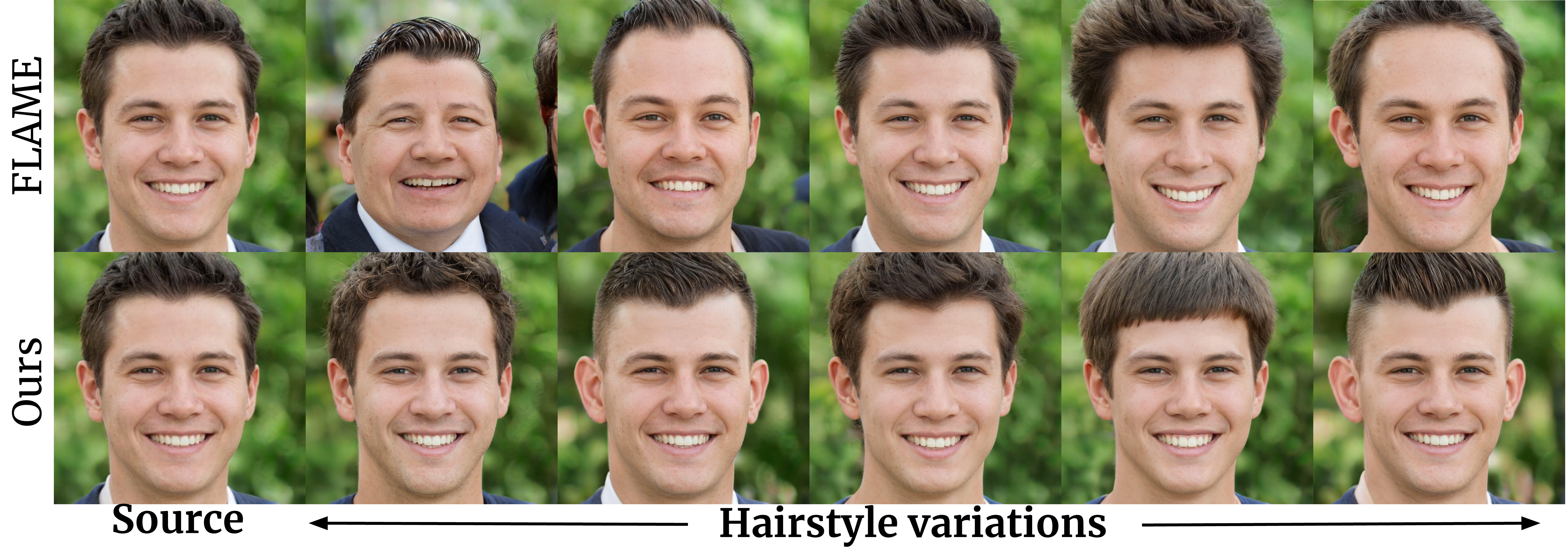}
    \vspace{-5mm}
    \caption{Qualitative comparison for diverse hairstyle editing. Observe the superior level of hairstyle variations and identity preservation from our method.}
    \label{fig:8_qual_compare}
    \vspace{-2mm}
\end{figure}

\textbf{Comparison with Infinite Direction methods.} We present qualitative results for the baseline method, FLAME, and our proposed method for diverse hairstyle generations in Fig.~\ref{fig:8_qual_compare}. We can observe that the baseline method changes the subject's identity while having limited style variations. On the other hand, FLAME can generate variations, but the identity and facial expressions of the subject are altered. The proposed method can generate diverse hairstyle variations like curls, texture, and bangs while preserving the subject's identity and retaining the other attributes. 

\noindent

\section{Comparison with single-direction based editing methods}
\label{sec:single_dir_compare}
\noindent 
We compare edits generated by our method against three state-of-the-art single-direction based editing methods: InterfaceGAN~\cite{shen2020interpreting}, StyleCLIP~\cite{patashnik2021styleclip}, StyleFlow~\cite{abdal2021styleflow} and CLIP2StyleGAN~\cite{abdal2022clip2stylegan}. Note that these methods can generate a single edit w.r.t. an attribute for a given image, whereas our method is trained to generate multiple edits w.r.t. an attribute. 
We performed editing to generate a single output for eyeglasses, smile, and age attributes using all these methods. We present the comparison results in Fig.~\ref{fig:s4_qual_edit_compare}. For CLIP2StyleGAN, we have generated results for only smile and eyeglasses attributes as CLIP2StyleGAN could not find disentangled edit directions for age. Our method achieves edits with high realism and disentanglement with superior identity preservation ability even with a single-directional edit. StyleFLow also achieves good edits with attribute disentanglement. However, StyleFlow requires additional attribute classifiers to obtain attribute scores which are required as input to edit any new input image. In Fig. \ref{fig:10_singe_direction_compare}
we compare our method against diverse edits using single edit direction methods. The variations in CLIP2StyleGAN and StyleFlow are generated by modifying the strength of the edit direction.

\begin{figure}[h]
    \centering    \includegraphics[width=0.7\linewidth]{figs_aaai/asm-aaai-supply-fig4-single-edit-compare.pdf}
    \caption{\textbf{Qualitative comparison with single direction methods.} StyleCLIP and CLIP2StyleGAN change the identity in age and eyeglass edits, respectively. InterFaceGAN entangles hair color with eyeglass attribute edits. StyleFlow changed the hair color while age editing. The proposed method can generate realistic edits without altering other attributes.}
    \label{fig:s4_qual_edit_compare}
\end{figure}

\begin{figure}[h]
    \centering    
    \includegraphics[width=1.0\linewidth]{figs_aaai/asm-aaai-fig4-single-direction-compare-small.pdf}
    \caption{Comparison for diverse attribute edits with single direction based editing methods - CLIP2StyleGAN~\cite{abdal2022clip2stylegan} and StyleFlow~\cite{abdal2021styleflow}. The variations in CLIP2StyleGAN and StyleFlow are generated by increasing the edit strength.}
\label{fig:10_singe_direction_compare}
\end{figure} 

\begin{figure*}[]
    \centering  \includegraphics[width=1.0\linewidth]{figs_aaai/asm-aaai-supply-diverse-hair-edits-2d-synthetic.pdf}
    \caption{Qualitative comparison with multi-direction based editing method - LatentCLR~\cite{latentclr} and StyleSpace~\cite{stylespace} for hairstyle variations. StyleSpace has $8$ style codes controlling hairstyles, whereas LatentCLR has $10$ directions in $\mathcal{W}$ space, which alter hairs. To generate more variations, we modify the strength of edit directions for all the methods.   
    StyleSpace can generate some hairstyle variations, but most hairstyles are highly similar (similar hairstyles indicated by the same colored bounding box). In contrast, our method generates highly diverse hairstyles, including bangs, curls, short hairs, and further variations among those. For LatentCLR, we could not find disentangled directions, and most of the directions entangle hair variations with age, skin color, and identity of the subject. This supports the intuition to model the rich distribution of edit directions with a generative model beyond selecting finite edit directions per attribute.}
    \label{fig:s5_qual_comparision_multi_dir}
\end{figure*}  

\section{Dataset Details}
\noindent

\begin{figure}[h]
    \centering    \includegraphics[width=0.7\linewidth]{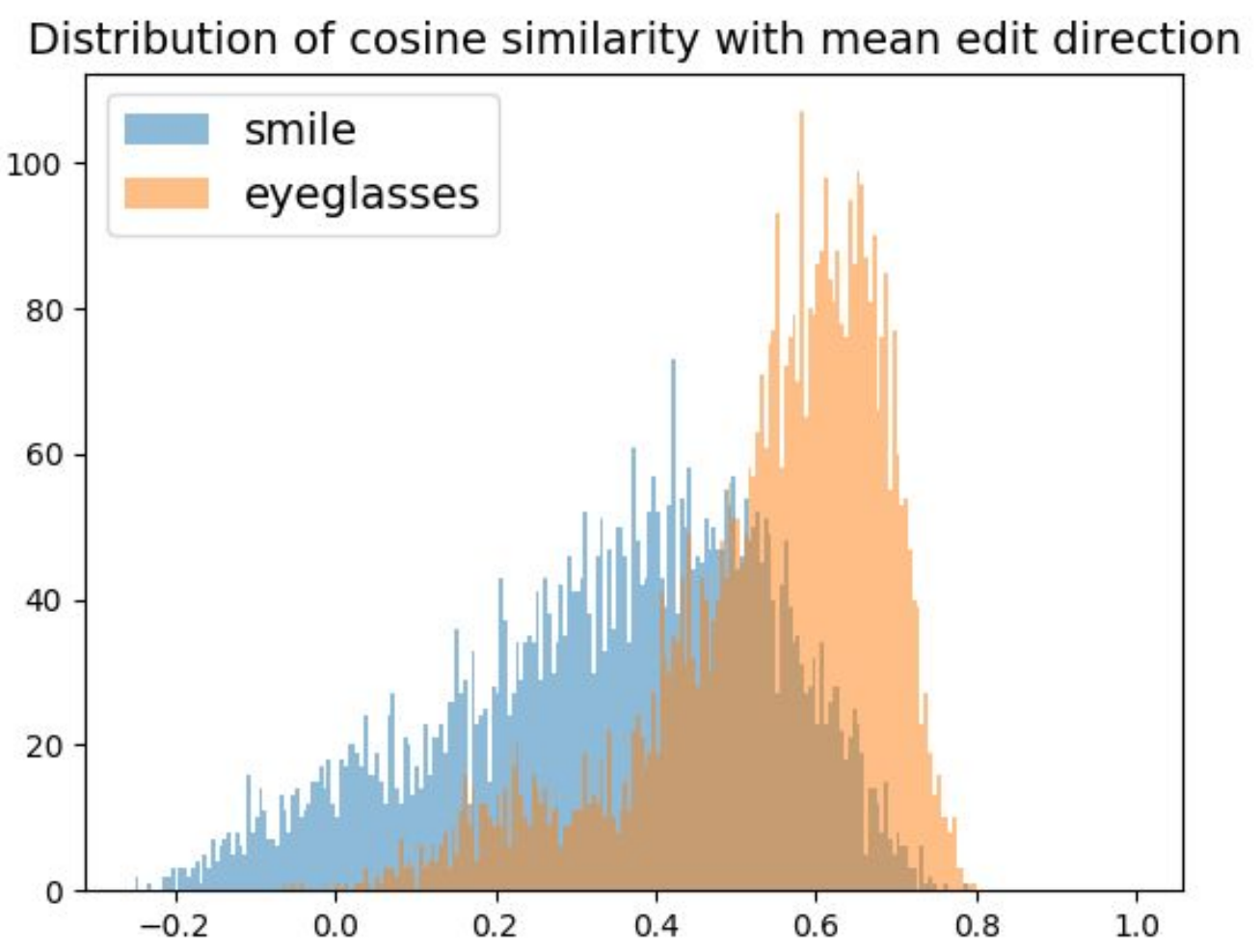}
    \caption{We plot a histogram of the Cosine Similarity of all edit directions with the mean direction. The spread of values suggests that the editing directions, although for the same attribute, showcase a large variety.}
    \label{fig:2_9_dataset}
\end{figure}

We create a synthetic dataset of image pairs with a single attribute change to learn the manifold of diverse attributes. To obtain such a dataset of paired samples, we employed two approaches: 1) Using existing single-direction editing method such as StyleCLIP~\cite{patashnik2021styleclip} to edit an input image for a given attribute; 2) Using the ``cut-paste" augmentation proposed in FLAME~\cite{parihar2022everything}. Our method performs good quality diverse edits when trained on the paired dataset obtained by both approaches, suggesting that our method is agnostic to the pair creation method. Next, we will provide details of these two approaches with examples. 

\vspace{4mm} 
\noindent \textbf{1. Existing single-direction editing method.}
We trained StyleCLIP~\cite{patashnik2021styleclip} mapper with text prompts ``bangs hairstyle", ``mohawk hairstyle", ``curly hairs", ``afro hairstyle", ``buzz cut", and ``bob cut" for hairstyles. We use the trained StyleCLIP models to perform single edits on a subset of CelebA-HQ~\cite{karras2017progressive} dataset. Examples of source and edited images are shown in Fig.~\ref{fig:dataset_pairs}. We create a dataset of $30K$ image pairs (edited and source) for each text prompt. The combined dataset of hairstyles and beards has $160K$ synthetic image pairs each. For the age attribute, we used SAM~\cite{sam_age}, a state-of-the-art age editing method, to generate $30K$ image pairs similarly. For car edits, we used StyleCLIP models trained with text prompts ``sports car" and ``classic car". To obtain image pairs for multiple church styles, we used an off-the-shelf style transfer network~\cite{park2020swapping}.

\begin{figure}[]
    \centering    \includegraphics[width=1.0\linewidth]{figs/asm-iccv-supply-fig5-dataset_pairs.pdf}
    \caption{\textbf{Examples of synthetic image pairs.}(Top) We present positive and negative image pairs generated by cut-pasting the attribute region (as explained in ~\ref{sec:cut-paste}). The augmented positive image is passed through the encoder and StyleGAN2 to obtain smooth inversion. (Bottom) Example image pairs generated by single direction editing method StyleCLIP, given the text prompts "bangs hairstyle", "mohawk hairstyle" and "curly hairs."  Note that the latent encodings of the negative and positive images are used to obtain the edit directions. 
    }
    \label{fig:dataset_pairs}
\end{figure}

\noindent \textbf{2. Cut-paste augmentation.}\label{sec:cut-paste}
For generating images for smile and eyeglass attributes, we used cut-paste augmentation as proposed in ~\cite{parihar2022everything}. Specifically, we sample a set of ``positive" images $X_p$ with attribute $a$ from the CelebA-HQ dataset using the attribute annotations from CelebA. We sample a set of ``negative" images $X_n$ that does not have the attribute $a$. Then we sample an image from $x_n \in X_n$ and $x_p \in X_p$ and mask the region of interest/part of the face that contains $a$ from $x_p$ and paste it onto $x_n$. For example, for the smile attribute, we used the mouth region; while for eyeglasses, we used eyeglass regions to perform cut-paste augmentation. These augmented images, when passed through the e4e encoder and StyleGAN2 generator, are smoothly blended due to smoothness prior to StyleGAN generator~\cite{chai2021using}. To obtain part segmentation masks, we use a few shot segmentation network~\cite{tritrong2021repurposing}, which uses StyleGAN features to perform segmentation. Specifically, we used five ground truth segmentations from CelebAHQ-Mask~\cite{CelebAMask-HQ} to train the few shot segmentation models. Some example image pairs generated by this operation are shown in Fig.~\ref{fig:dataset_pairs}.

\noindent \textbf{How diverse are the edit directions in the dataset?}\label{sec:diversity_of_dataset}
We show the distribution of cosine similarity of the edit directions with the mean edit direction in Fig.~\ref{fig:2_9_dataset}. Observe that the obtained directions have high diversity and do not align with the mean direction.

\section{Model architecture} 
\noindent
We implemented our denoising network as an MLP network with $10$ fully connected layers with $2048$ neurons in each layer. Additionally, we added a time conditioning by first encoding the timestamp with $128$ dimensional positional encoding. The encoded time embeddings are passed through another MLP network of two fully connected layers with $256$ neurons each. Time conditioning is added to all hidden layers of the base MLP network. A skip connection is added after each linear layer, followed by a layer norm layer. Layer norm is not added in the final layer.

\section{Computational requirement}
We performed all our experiments on a single NVIDIA-A5000 GPU. The training time of DDPM model on a dataset for latent directions is 1 hour on a single GPU for a batch size of 64, although 3-4 such models can be trained on a single A5000 at a time since the runs require less memory. 

\section{Additional Results} 
\label{sec:addn_results}
\noindent

\textbf{Sequential editing of multiple attributes.} As our method generates global edit directions, we can combine edit directions for multiple attributes to obtain a composite edit. We present results for composite editing of three attributes - smile, hairstyle, and eyeglasses, in Fig.~\ref{fig:11_multi_attrs}. For each attribute, we randomly sample an edit direction from the trained diffusion model and add it to the latent code of the source image to perform sequential editing. Observe that our method can generate consistent edits that combine multiple diverse attributes. 

\begin{figure}[h]
    \centering    \includegraphics[width=0.85\linewidth]{figs/asm-iccv-fig11-seq-edits.pdf}
    \vspace{-2mm}
    \caption{We sequentially perform smile, hairstyle, and eyeglasses edits with two variations per attribute.}
    \vspace{-4mm}
    \label{fig:11_multi_attrs}
\end{figure}

\begin{figure}[h]
    \centering    \includegraphics[width=\linewidth]{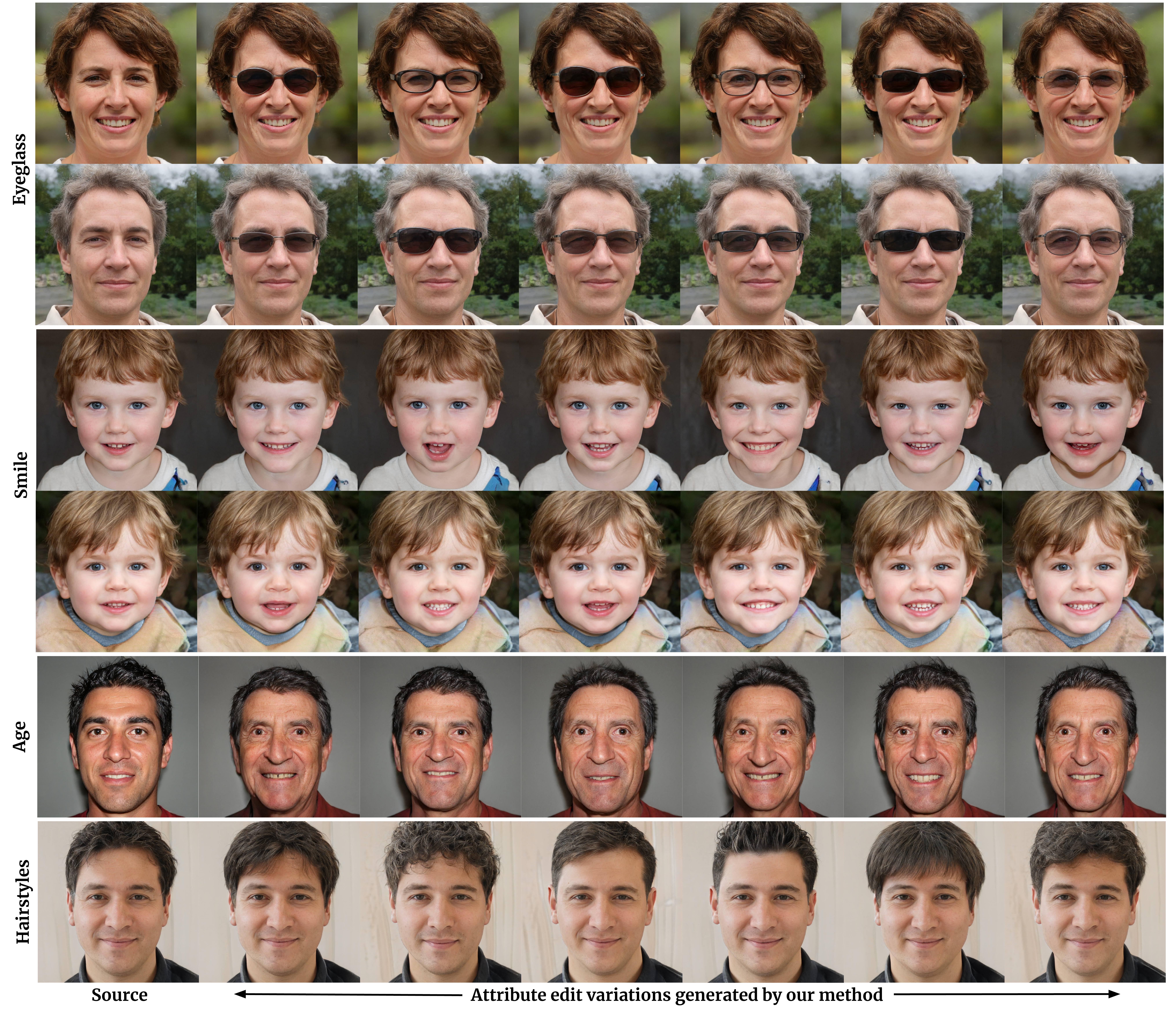}
    \vspace{-2mm}
    \caption{Diverse attribute edits for age, smile, eyeglass, and hairstyles on face images.}
    \vspace{-4mm}
    \label{fig:s7_addn_results_face}
\end{figure}

We present additional results for diverse attribute editing on real and synthetic images in Fig. \ref{fig:s7_addn_results_real_faces} and Fig. \ref{fig:s7_addn_results_face}, respectively. For real images, we first invert the input image into $\mathcal{W+}$ latent space using e4e~\cite{e4e} encoder model. The proposed method can generate diverse and realistic attribute variations for both real and synthetic datasets. Additionally, we present results for diverse editing on car styles and church styles in Fig. \ref{fig:s7_addn_results_church}.

\begin{figure}[h]
    \centering    \includegraphics[width=\linewidth]{figs_aaai/asm-aaai-supply-fig8-church-edits.pdf}
    \vspace{-2mm}
    \caption{Diverse attribute edits for car styles - sports car and classic car and church styles.}
    \vspace{-4mm}
    \label{fig:s7_addn_results_church}
\end{figure}

\begin{figure}[h]
    \centering    \includegraphics[width=\linewidth]{figs/asm-iccv-supply-fig7-real_img_edits.pdf}
    \vspace{-2mm}
    \caption{\textbf{Diverse attribute edits on real images.} Given a real image, we encode it using e4e~\cite{e4e} encoder and perform diverse attribute editing on the obtained latent code.}
    \vspace{-4mm}
    \label{fig:s7_addn_results_real_faces}
\end{figure}

\section{Editing on 3D faces}
\noindent
We present diverse attribute editing results on 3D aware GAN, EG3D~\cite{eg3D}. The proposed method can generate diverse attribute edits that are 3D consistent and preserve the identity. We present results for editing in 3D along with the surface maps of the edited geometry in Fig.~\ref{fig:eg3d_edits_supply}. Observe that the shape of eyeglasses and hairstyles are changed with diverse edits as visible in the surface maps. We leave a detailed exploration of other 3D GANs as a future work. 

\begin{figure*}[h]
    \centering    \includegraphics[width=0.8\linewidth]{figs_aaai/asm-aaai-supply-eg3d-vis.pdf}
    \caption{\textbf{Diverse attribute variations for 3D aware GANs.} Our method can generate diverse edits which actually alter the 3D geometry of the face, as shown in the surface maps.}
\label{fig:eg3d_edits_supply}
\end{figure*}

 \clearpage
\bibliographystyle{unsrtnat}
\bibliography{neuripsw}